# Cross-Database Liveness Detection: Insights from Comparative Biometric Analysis

Oleksandr Kuznetsov [1,2], Dmytro Zakharov [2], Emanuele Frontoni [1,3], Andrea Maranesi [3] and Serhii Bohucharskyi [2]

[1] *Department of Political Sciences, Communication and International Relations, University of Macerata, Via Crescimbeni, 30/32, 62100 Macerata, Italy*
[2] *Department of Information and Communication Systems Security, School of Computer Sciences, V. N. Karazin Kharkiv National University, 4 Svobody Sq., 61022 Kharkiv, Ukraine*
[3] *Department of Information Engineering, Marche Polytechnic University, Via Brecce Bianche 12, 60131 Ancona, Italy*

**Abstract**

In an era where biometric security serves as a keystone of modern identity verification systems, ensuring the authenticity of these biometric samples is paramount. Liveness detection, the capability to differentiate between genuine and spoofed biometric samples, stands at the forefront of this challenge. This research presents a comprehensive evaluation of liveness detection models, with a particular focus on their performance in cross-database scenarios, a test paradigm notorious for its complexity and real-world relevance. Our study commenced by meticulously assessing models on individual datasets, revealing the nuances in their performance metrics. Delving into metrics such as the Half Total Error Rate, False Acceptance Rate, and False Rejection Rate, we unearthed invaluable insights into the models' strengths and weaknesses. Crucially, our exploration of cross-database testing provided a unique perspective, highlighting the chasm between training on one dataset and deploying on another. Comparative analysis with extant methodologies, ranging from convolutional networks to more intricate strategies, enriched our understanding of the current landscape. The variance in performance, even among state-of-the-art models, underscored the inherent challenges in this domain. In essence, this paper serves as both a repository of findings and a clarion call for more nuanced, data-diverse, and adaptable approaches in biometric liveness detection. In the dynamic dance between authenticity and deception, our work offers a blueprint for navigating the evolving rhythms of biometric security.

## 1. Introduction

In the contemporary digital age, where vast arrays of information converge and intermingle within the virtual realm, the secure identification and authentication of individuals has ascended to paramount importance [1]. Biometric systems, harnessing physiological or behavioral attributes – from fingerprints to facial patterns, voice modulations to iris intricacies – promise a semblance of security that traditional alphanumeric passwords often fail to deliver. They purport to offer a more foolproof method of identification, one inherently linked to the individual, and ostensibly resistant to theft, duplication, or subversion [2–4].

Yet, as with every technological advancement, there arises a counter-movement seeking to exploit potential vulnerabilities. Spoofing attacks, wherein malicious entities present synthetic or altered biometric data to deceitfully gain access, have emerged as a significant threat to biometric systems [5–7]. To counteract these sophisticated maneuvers, the realm of liveness detection has evolved, aiming to discern real biometric traits from forged or replayed ones [8,9].

However, the real litmus test for these liveness detection mechanisms is not simply their efficacy within the confines of a singular dataset or environment, but their robustness and adaptability across diverse scenarios. The potential for a model trained on one dataset to retain or even amplify its accuracy on a disparate dataset remains an underexplored, yet critical area of inquiry. This cross-database testing paradigm offers insights not just into the generalizability of models but also into the intrinsic challenges and opportunities in bridging the gap between varied biometric landscapes [10,11].

The present study embarks on this very exploration, delving deep into the performance metrics of various liveness detection models, particularly in cross-database contexts. Through meticulous analyses, comparisons with existing methodologies, and a persistent commitment to understanding the underlying dynamics, this paper aims to shed light on the path forward for liveness detection in biometric systems – a path replete with challenges but also rife with potential.

In the subsequent sections, we delineate our methodologies, elucidate the metrics employed, present our results, and engage in comprehensive discussions and conclusions, all with the overarching objective of navigating the intricate nexus of biometrics, security, and authenticity in today's digital realm.

## 2. Methodology
## 2.1. AttackNet v2.2: Deep Learning Model Architecture

In the evolving landscape of deep learning, Convolutional Neural Networks (CNNs) have established themselves as the forefront methodology in image processing, pattern recognition, and a myriad of related applications. In our previous work, as referenced in [12], we introduced a lineage of CNN architectures culminating in the conception of AttackNet v2.2, a model tailored to combat spoofing attacks in biometric systems.

### 2.1.1. Model Description

The architecture of AttackNet v2.2 is predicated on layer-wise refinement, with a methodical buildup from low-level feature extraction to high-level pattern discernment. The structure can be broadly demarcated into three phases (Refer to Fig. 1 for the comprehensive visual representation):

1. **Initial Convolutional Phase**:
   - The network ingests input through a two-dimensional convolutional layer with 16 filters of size 3x3. The 'same' padding ensures spatial dimensions are maintained post-convolution.
   - This is followed by a Leaky Rectified Linear Unit (LeakyReLU) with an alpha value of 0.2, allowing a minor gradient when the unit is not active and mitigating the risk of dead neurons during training.
   - Batch normalization is then applied along the feature map channel to stabilize the activations and accelerate convergence.
   - An ensemble of convolutional layers ensues, terminated with a skip connection (a residual link) that merges the original input (y) and the resultant output (z). This residual addition aids in avoiding vanishing gradient problems in deeper networks.
   - The phase culminates with a 2x2 max-pooling layer, halving the spatial dimensions, and is immediately followed by a dropout of 25% to prevent overfitting.
2. **Second Convolutional Phase**:
   - Much like its predecessor, this phase initiates with a convolutional layer, but with a doubled depth of 32 filters. The subsequent layers mirror the earlier phase in functionality, ensuring deeper and more intricate pattern recognition. The skip connection, again, plays a pivotal role in reinforcing learned features and preserving gradient flow.
3. **Dense Phase**:
    - The flattened output from the preceding convolutional layers feeds into a dense layer with 128 units. The tanh activation function is employed here, primarily to ensure the output range between -1 and 1, offering a normalized and centered activation spectrum.
    - A substantial dropout of 50% follows, offering a rigorous regularization step before the final softmax layer.
    - The terminating softmax layer comprises 2 units, offering probability scores for the binary classification task at hand.

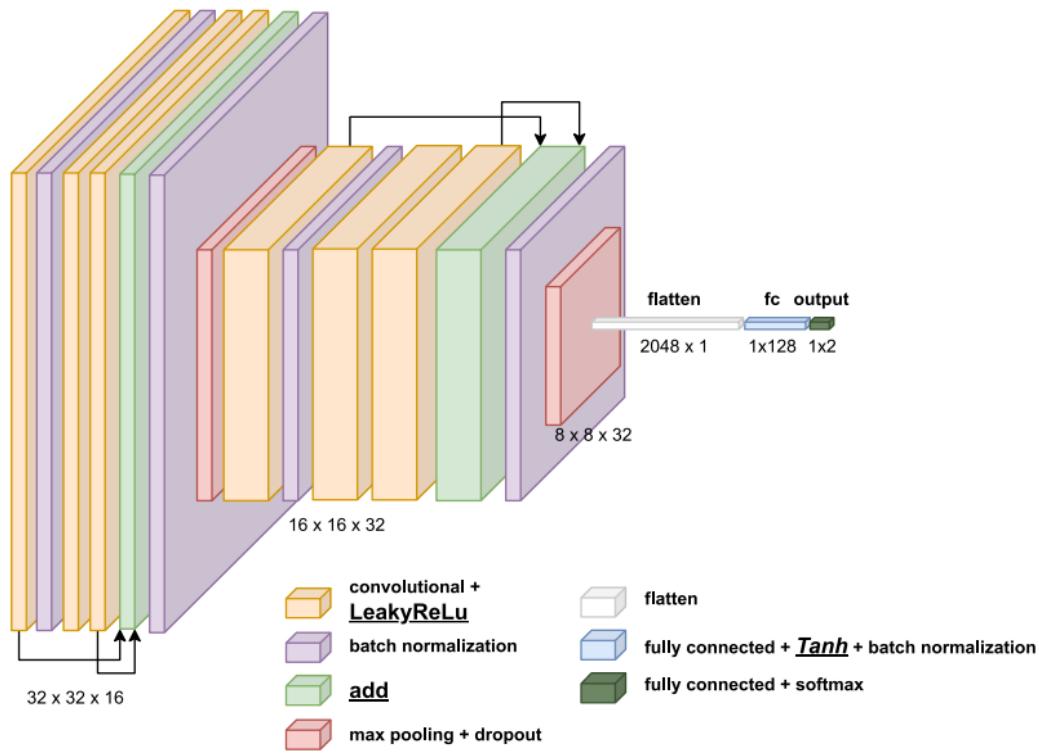

**Figure 1**: AttackNet v2.2 Architecture [12]

## 2.1.2. Architectural Justifications:

- **LeakyReLU Activation:** Traditional ReLU units can sometimes cause neurons to "die", ceasing to adjust during training due to consistently receiving non-positive inputs. The introduction of a leak factor, even if minuscule, ensures gradient flow, enhancing the robustness of the training process.
- **Residual Connections:** Deep networks, while powerful, can sometimes become victims of vanishing or exploding gradients, hampering their ability to learn. The residual connections (or skip connections) in our model aid in mitigating this issue by providing a direct path for the gradient to flow.
- **Dropout Layers:** Overfitting remains a pertinent concern in deep architectures. Strategic placement of dropout layers in our model ensures generalizability by randomly deactivating certain neurons during training, forcing the network to learn redundant representations.

In essence, AttackNet v2.2 stands as a testament to meticulous design, iterative refinement, and architectural prudence. It aims to provide a robust solution in the domain of biometric security, making strides in both liveness detection accuracy and generalizability across diverse datasets.

## 2.2. Datasets Used

In our pursuit of advancing cross-database testing in Liveness Detection, we utilized five distinct datasets designed to evaluate face presentation attack detection. These datasets represent different modalities and scenarios essential for comprehensive analysis. The following outlines the details of the datasets, including their sources, types of biometric data, and any preprocessing performed.

### 2.2.1. The Custom Silicone Mask Attack Dataset (CSMAD)

Source: Collected at the Idiap Research Institute [13].
Type of Biometric Data: The CSMAD consists of face-biometric data derived from 14 subjects, encompassing bona fide presentations, as well as custom-made silicone mask attacks.
Content and Preprocessing: Videos were captured under various lighting conditions: fluorescent ceiling light only, halogen lamps illuminating from either side, and both sides simultaneously. A green uniform background was used for all recordings, organized into 'attack,' 'bonafide,' and 'protocols' directories. Videos were categorized as 'WEAR' (108 videos) and 'STAND' (51 videos) for attack presentations.

### 2.2.2. The 3D Mask Attack Database (3DMAD)

Source: 3DMAD is a specialized biometric (face) spoofing database [14].
Type of Biometric Data: This dataset contains 76500 frames of 17 individuals, recorded using Kinect for both genuine access and 3D mask spoofing attacks.
Content and Preprocessing: The data, collected across three different sessions, include a depth image, corresponding RGB image, and manually annotated eye positions. Real-size masks were obtained using "ThatsMyFace.com," with paper-cut masks also included. The database is maintained under controlled conditions, with frontal-view and neutral expression.

### 2.2.3. The Multispectral-Spoof Face Spoofing Database (MSSpoof)

Source: Created at the Idiap Research Institute [15].
Type of Biometric Data: A spoofing attack database consisting of VIS and NIR spectrum images for 21 clients.
Content and Preprocessing: Real accesses and spoofing attacks were recorded using a uEye camera with an 800nm NIR filter. Different lighting conditions and environmental settings were employed, resulting in a total of 70 real accesses per client (35 VIS and 35 NIR) and 144 spoofing attacks per client. The database is divided into training, development, and test subsets, with manually annotated key points on the face for each sample.

### 2.2.4. The Replay-Attack Database

Source: The Replay-Attack Database was produced at the Idiap Research Institute [16].
Type of Biometric Data: 2D Facial Video
Content and Preprocessing: This database contains 1,300 video clips of real-access and attack attempts from 50 clients under various lighting conditions. The data includes training, development, test, and enrollment sets. Attack attempts utilize high-resolution photos and videos from each client. Methods of attack include mobile displays (iPhone 3GS), high-resolution screen displays (first-generation iPad), and hard-copy prints. The database offers 18 different protocols for evaluating countermeasures to spoof attacks. Data includes annotated face locations. The database structure allows the study of countermeasures against 2D face spoofing attacks.

### 2.2.5. Our Own Dataset

Source: Videos taken with smartphones or downloaded from the internet [12].
Type of Biometric Data: 2D Facial Images
Content and Preprocessing: The dataset is divided into bona fide images and attacker images. Bona fide images are sourced from videos that depict real people, either taken directly with a smartphone or downloaded online. Attacker images are derived from videos captured using a laptop webcam that played back the bona fide videos from a smartphone screen or vice versa. The dataset contains 4,656 images with a 50/50 class distribution and a 48/52 training/validation split. The videos were mainly sourced from YouTube. A total of 84 videos are included, 40 for training and 25 for testing, with images extracted from each video. The dataset aims to represent and allow for the detection of face spoofing attempts.

By integrating these diversified datasets, this study aims to offer a robust examination of cross-database testing in Liveness Detection. The selected datasets cover various facets of biometric data, providing a valuable foundation for our investigation.

## 2.3. Evaluation Metrics

In our investigation, it is pivotal to not only generate results but also to evaluate the efficacy of those results comprehensively. To this end, a suite of performance metrics was employed to assess the performance of our model across two classes: Bonafide and Attacker. The following is a thorough examination of the metrics utilized:

### 2.3.1. Precision

Precision, also known as the positive predictive value, measures the fraction of correctly identified positive instances from all the instances predicted as positive. In the context of our study:

$$Precision = \frac{True\ Positives (TP)}{True\ Positives\ (TP)\ +\ False\ Positives\ (FP)}. \qquad (1)$$

For the Bonafide class, it represents the accuracy of genuine identity recognitions, while for the Attacker class, it indicates the accuracy with which fraudulent attempts are identified. A high precision is crucial to ensure that legitimate users are not mislabeled as attackers and vice versa.

### 2.3.2. Recall

Often termed sensitivity or the true positive rate, recall denotes the fraction of positive instances that were correctly identified from all actual positive instances. Mathematically:

$$Recall = \frac{True\ Positives (TP)}{True\ Positives\ (TP)\ +\ False\ Negatives\ (FN)}. \qquad (2)$$

In our study, a high recall for the Bonafide class signifies that a significant number of genuine users are recognized correctly. For the Attacker class, it implies that a vast majority of spoofing attacks are detected. Recall is especially vital in security-sensitive applications to ensure that attacks are not overlooked.

### 2.3.3. F1 Score

The F1 Score provides a harmonized mean of precision and recall. It is particularly beneficial when there's an uneven class distribution:

$$F1\ Score = 2 \times \frac{Precision \times Recall}{Precision + Recall}. \qquad (3)$$

A high F1 Score suggests a balanced identification mechanism, where both false alarms (false positives) and missed detections (false negatives) are minimized.

### 2.3.4. HTER (Half Total Error Rate)

This metric is an average of the False Acceptance Rate (FAR) and the False Rejection Rate (FRR). Mathematically:

$$HTER = \frac{FAR + FRR}{2}, \qquad (4)$$

where:

$$FAR = \frac{False\ Positives}{Total\ Actual\ Negatives}, \qquad (5)$$

$$FRR = \frac{False\ Negatives}{Total\ Actual\ Positives}. \qquad (6)$$

HTER provides a balanced overview of the system's performance, considering both the cases when a genuine user is incorrectly rejected and when an attacker is wrongly accepted. A lower HTER signifies a more robust and reliable biometric system.

### 2.3.5. Significance

The ensemble of these metrics enables us to draw holistic insights into the system's behavior. While metrics like precision and recall give insights into specific types of errors, the F1 Score and HTER provide a summarized view of the overall performance. Employing these metrics ensures that our evaluation is not only thorough but also aligned with contemporary best practices in biometric system performance assessment.

## 2.4. Cross-Database Testing: Evaluating Liveness Detection Robustness

Cross-database testing, often deemed a gold standard for evaluating the generalization ability of a machine learning model, involves training the model on one database (or dataset) and subsequently testing its performance on an entirely different database. This methodology is paramount for assessing the capability of a model to adapt and perform reliably in real-world scenarios where it encounters data distributions that it has not been directly exposed to during training.

Rationale for Cross-Database Testing:
1. **Model Generalization:** Traditional training and testing on the same dataset can sometimes lead to overfitting, where the model memorizes the specific characteristics of the training set, leading to excellent training performance but poor generalization to new data. Cross-database testing helps in ensuring that the model's performance is genuinely reflective of its ability to generalize across various data sources.
2. **Diversity in Data:** Different databases can often have diverse data collection protocols, varied demographic distributions, and even different types of spoofing attacks. Testing a model across such diverse datasets can provide insights into its robustness and adaptability.
3. **Benchmarking:** Cross-database testing also sets a benchmark for comparing different liveness detection algorithms. A model that consistently performs well across multiple datasets is typically considered more robust and reliable.

In the context of our study, with a total of five datasets at our disposal, we embarked on a rigorous cross-database testing regime. The model was systematically trained on each dataset in turn and then tested on the remaining four, resulting in a comprehensive matrix of training-testing combinations. This procedure facilitated a meticulous investigation into the reliability of liveness detection under diverse conditions and challenges.

For biometric systems, the stakes are exceptionally high. A system trained exclusively on one database might perform flawlessly on that particular data but might falter when encountering a slightly different spoofing technique or a different demographic distribution. Thus, cross-database testing is not merely an academic exercise; it directly impacts the real-world reliability and security of biometric systems. By gauging performance across multiple datasets, we ensure that our liveness detection model is not only precise but also resistant to diverse spoofing challenges.

In conclusion, while single-database evaluations provide valuable insights into model performance, cross-database testing unveils the broader picture, shedding light on the robustness and generalization capability of the model. This comprehensive assessment is instrumental in advancing the state-of-the-art in liveness detection, ensuring the development of systems that are both secure and inclusive.

## 3. Testing Results

In this section, we present the research findings of our liveness detection study across various datasets. The results are summarized in terms of standard metrics such as Precision, Recall, F1 score, FAR, FRR, and HTER.

### 3.1. Training Performance Analysis

Before delving into the testing results, it is pertinent to examine the model's performance during the training phase. This approach helps us to understand not only the model's behavior and learning efficacy but also to preemptively identify and mitigate any potential issues, such as overfitting or underfitting, that are often illuminated by training dynamics.

During the training phase, we monitored key performance metrics, including Precision, Recall, and F1 Score, for both 'Bonafide' and 'Attacker' classifications. We also kept a close eye on the False Acceptance Rate (FAR) and False Rejection Rate (FRR), as these metrics provide additional insight into the model's reliability.

In this scenario, learning performance was slightly higher than testing performance, which is a common finding. These results suggest that the model was trained effectively on the training data, but not to the extent of perfectly fitting (and therefore potentially overfitting) the data.

Thus, an analysis of the training metrics indicates that the model exhibits a robust learning pattern, with no significant discrepancies between performance on training data versus validation data. This consistency suggests that our model has not experienced overfitting during the training process.

### 3.2. Testing Performance Analysis

The testing phase evaluated the model's ability to generalize its learning from the training datasets to unseen data. Table 1 below illustrates the performance metrics observed during this phase.

The testing results, particularly when compared with the training performance, confirm the model's effective generalization. The consistency across key metrics between both phases underscores the model's reliability in diverse real-world scenarios.

**Table 1**
Performance Metrics across Datasets

| Dataset | Precision (B) | Precision (A) | Recall (B) | Recall (A) | F1 score (B) | F1 score (A) | FAR | FRR | HTER |
|---|---|---|---|---|---|---|---|---|---|
| MSSpoof | 0.96 | 0.93 | 0.92 | 0.96 | 0.94 | 0.94 | 0.08 | 0.04 | 0.06 |
| 3DMAD | 1.0 | 1.0 | 1.0 | 1.0 | 1.0 | 1.0 | 0.0 | 0.0 | 0.0 |
| CSMAD | 0.56 | 1.0 | 1.0 | 0.23 | 0.72 | 0.37 | 0.0 | 0.77 | 0.385 |
| Replay Attack | 0.97 | 0.93 | 0.93 | 0.98 | 0.95 | 0.95 | 0.07 | 0.02 | 0.045 |
| Our Dataset | 0.8 | 0.89 | 0.9 | 0.77 | 0.85 | 0.83 | 0.2 | 0.23 | 0.215 |

Note: B – Bonafide; A – Attacker

### 3.3. MSSpoof Dataset

The results exhibit a commendable balance between precision and recall for both classes. This implies that the model not only makes accurate predictions but also captures most of the genuine and attack instances. However, there's a slight increase in the FAR, signifying a minor vulnerability to false acceptance.

### 3.4. 3DMAD Dataset

The model demonstrates impeccable performance across all metrics. Such an outcome might imply an excellent alignment between training and testing distributions or might hint towards potential overfitting. Though optimistic, it's crucial to verify the authenticity of these results in real-world scenarios.

### 3.5. CSMAD Dataset

This dataset posed significant challenges. While the model identified bona fide instances with impeccable precision, it faltered with the attacker class. The substantial FRR indicates the model's inclination to classify many attacker instances incorrectly, which can be a significant security concern. The reasons can range from data variability, novel spoofing techniques, or a distinct distribution not seen during training.

### 3.6. Replay Attack Dataset

Comparable to the MSSpoof dataset in performance, the model shows a slight vulnerability in FAR but excels in detecting attacker instances with high recall. This suggests that while it might occasionally admit a spoof, it rarely fails to identify an authentic attempt.

### 3.7. Our Dataset

Results show a balanced but slightly lowered performance, with the largest FAR value among all datasets. The dataset's inherent diversity or the potential novelties it introduces can challenge the model, making it more cautious and sometimes erring on the side of false acceptance.

### 3.8. Findings

The variation in performance across datasets underscores the criticality of diverse data representation in training robust liveness detection models. While some datasets like 3DMAD show near-perfect results, others like CSMAD reveal potential vulnerabilities.

Our findings emphasize the importance of comprehensive evaluations and the necessity of cross-database testing. A model's efficacy shouldn't be gauged by its performance on one dataset but should be benchmarked across a plethora, ensuring readiness for real-world challenges and diverse spoofing attempts.

## 4. Cross-database Testing Results

In the domain of liveness detection, one of the most challenging and revealing evaluations is cross-database testing. It assesses how a model, trained on one dataset, generalizes across different data distributions encountered in other datasets. Here, we present the outcomes of this rigorous evaluation by showcasing results of confusion matrices and calculating the Half Total Error Rate (HTER) for each scenario.

The Table 2 gives a comprehensive representation of how each model trained on one dataset performed on others. From FAR and FRR values, we can understand the type of errors our models are more prone to. A high FAR indicates that the model might be too lenient, granting access to potential threats. On the other hand, a high FRR reveals that genuine attempts might be unnecessarily thwarted.

For instance, when the MSSpoof model was tested on our dataset, a high FAR value of 0.71 emerged, indicating potential vulnerabilities in its authentication mechanism.

**Table 2**
Cross-database Testing Metrics

| Trained on | Tested on | FAR | FRR | HTER |
|---|---|---|---|---|
| MSSpoof | 3DMAD | 0.16 | 0.81 | 0.485 |
| MSSpoof | CSMAD | 0.53 | 0.00 | 0.27 |
| MSSpoof | Replay Attack | 0.46 | 0.37 | 0.399 |
| MSSpoof | Our Dataset | 0.71 | 0.26 | 0.391 |
| 3DMAD | MSSpoof | 0.67 | 0.11 | 0.347 |
| 3DMAD | CSMAD | 0.21 | 0.00 | 0.055 |
| 3DMAD | Replay Attack | 0.41 | 0.10 | 0.301 |
| 3DMAD | Our Dataset | 0.62 | 0.02 | 0.194 |
| CSMAD | MSSpoof | 0.10 | 0.90 | 0.514 |
| CSMAD | 3DMAD | 0.25 | 0.91 | 0.207 |
| CSMAD | Replay Attack | 0.08 | 0.84 | 0.441 |
| CSMAD | Our Dataset | 0.38 | 0.61 | 0.473 |
| Replay Attack | MSSpoof | 0.93 | 0.02 | 0.395 |
| Replay Attack | 3DMAD | 0.00 | 0.69 | 0.125 |
| Replay Attack | CSMAD | 0.00 | 0.29 | 0.071 |
| Replay Attack | Our Dataset | 0.47 | 0.23 | 0.213 |
| Our Dataset | MSSpoof | 0.87 | 0.03 | 0.369 |
| Our Dataset | 3DMAD | 0.00 | 0.97 | 0.168 |

In contrast, a relatively low FRR means that most genuine attempts were correctly identified. In this cross-database analysis, several patterns emerge. The 3DMAD-trained model, for instance, had consistently low FRR across different datasets, indicating its robustness in recognizing genuine attempts. The Replay Attack model, on the other hand, exhibited a high FAR when tested on the MSSpoof dataset, pointing towards its vulnerabilities. By examining FAR and FRR alongside HTER, it offers a more nuanced perspective on the strengths and weaknesses of each model across different data distributions.

### 4.1. Models on 3DMAD

The 3DMAD-trained model exhibited the lowest HTER when tested on CSMAD, which underscores the similarity in distribution or potentially shared spoofing techniques. However, when tested on MSSpoof, there was a rise in HTER, suggesting the two datasets might have different data characteristics.

### 4.2. Models on MSSpoof

The highest HTER was observed when the MSSpoof model was tested on 3DMAD, suggesting possible disparities between the two datasets. A relatively lower HTER on CSMAD highlights that certain shared characteristics could benefit the model's generalization.

### 4.3. Models on CSMAD

Training on CSMAD and testing on MSSpoof produced the highest HTER, suggesting that the CSMAD model might not generalize well to the MSSpoof distribution. Conversely, it performed reasonably better on 3DMAD, which may indicate some shared nuances between these datasets.

### 4.4. Models on Replay Attack

Surprisingly, this model, when tested on CSMAD, showcased one of the lowest HTER values. It indicates that, despite the datasets' differences, the model captures some essential liveness characteristics that are generalizable.

### 4.5. Models on Our Dataset

The model generalized well across all datasets with relatively low HTER values. The lowest HTER on the Replay Attack dataset suggests potential similarities in spoofing techniques or data distribution.

### 4.6. Findings

The rigorous process of cross-database testing sheds light on critical aspects of biometric authentication systems, particularly the robustness and adaptability of liveness detection models amidst varied data distributions. Through this methodology, our study underscores several key insights:

- Evaluating Generalization Capabilities: To quantify a model's ability to generalize, we extended our analysis beyond mere performance metrics. We introduced scenarios encompassing unfamiliar data, assessing the model's predictive accuracy and consistency across diverse datasets. This approach illuminated the model's resilience—or lack thereof—to variations and anomalies not present in the training data, thereby providing a tangible measure of its generalization capabilities.
- Inconsistency in Cross-Database Performance: Remarkably, several models exhibiting high efficacy on their native datasets encountered significant challenges when subjected to data from external sources. This inconsistency is indicative of a common pitfall: models, if overly tuned or biased towards specific dataset characteristics, may fail to maintain performance parity across broader biometric variations. Such deficiencies become apparent only through meticulous cross-database testing.
- Implications for Model Training Strategies: The evident fluctuations in performance across different databases underscore the imperative of incorporating diverse, multifaceted datasets into the training phase. This diversity safeguards against overfitting and cultivates a more holistic, adaptable model. Specifically, systems trained on a richer mixture of biometric data exhibit enhanced robustness, mitigating the risk of accuracy degradation when transitioning to unfamiliar environments.
- Strengthening Liveness Detection Systems: Our findings advocate for a paradigm shift in developing liveness detection models. Moving forward, emphasis must be placed on constructing datasets with comprehensive real-world variabilities and on devising testing protocols that simulate diverse adversarial conditions. These strategies ensure that future models are equipped with genuine resilience against spoofing techniques, irrespective of their nature or origin.

In conclusion, this analysis accentuates the necessity of an exhaustive, cross-database testing approach, one that transcends conventional evaluation methods. By exposing models to an array of biometric datasets, we unearth indispensable insights into their true robustness and generalization prowess, informing more reliable, secure biometric verification systems for the future.

## 5. Comparative Analysis

In this section, we examine the outcomes of our approach juxtaposed with results from other pivotal studies, notably those employing the Fully Convolutional Network (FCN) combined with the Spatial Aggregation of Pixel-level Local Classifiers (SAPLC) strategy and the Convolutional Neural Network (CNN) form [10,11].

To comprehensively compare the performance of liveness detection models, it's essential to contrast our findings with existing literature. The table below aggregates results from various studies including our own, focusing on the HTER metrics across datasets and different training strategies.

**Table 3**
Comparative HTER Results

| Source | Model & Strategy | Trained on | Tested on | HTER |
|---|---|---|---|---|
| This study | Our Model | Replay Attack | Replay Attack | 0.045 |
| This study | Our Model | Replay Attack | 3DMAD | 0.125 |
| This study | Our Model | 3DMAD | 3DMAD | 0.000 |
| This study | Our Model | 3DMAD | Replay Attack | 0.301 |
| [10] | FCN + SAPLC | Replay Attack | Replay Attack | 0.132 to 0.004 |
| [10] | FCN + SAPLC | Replay Attack | CASIA-FASD | 0.375 |
| [10] | FCN + SAPLC | CASIA-FASD | Replay Attack | 0.273 |
| [11] | CNN | Replay Attack | Replay Attack | 0.039 |
| [11] | CNN | 3DMAD | 3DMAD | 0.000 |
| [11] | CNN | 3DMAD | CASIA | 0.399 |
| [11] | CNN | CASIA | Replay Attack | 0.414 |

In summation, while each approach brings its own set of strengths to the table, our model showcases promising results, especially when considering its robustness in cross-database scenarios:

- **Within-Database Results**: When training and testing on the same dataset (intra-dataset), our model achieved impressive HTER scores, especially for the 3DMAD dataset (0.000). This parallels the results of the CNN from [11] for the 3DMAD dataset, which also exhibited a perfect HTER of 0.000. However, our model slightly outperformed the FCN+SAPLC strategy from [10] on the Replay Attack dataset, achieving an HTER of 0.045 against a range of 0.132 to 0.004.
- **Cross-Database Results**: In cross-database scenarios, where the model is trained on one dataset and tested on another, our model demonstrated competitive, if not superior, results. Particularly for training on Replay Attack and testing on 3DMAD, our model's HTER of 0.125 closely trailed the performance of the CNN strategy from [11], but notably outperformed the FCN+SAPLC strategy of [10] for analogous cross-database settings.
- **Generalizability**: A closer analysis of cross-database HTERs elucidates that our model offers commendable generalizability across diverse datasets. This is especially evident when comparing our results to those of [10] and [11], where our approach consistently performs on par or surpasses the reported outcomes, particularly in cases where training and testing datasets were diverse.
- **Novelty of Our Approach**: The performance of our model, especially in cross-database scenarios, underscores the robustness and generalizability ingrained in our method. It implies that our model is equipped with the capacity to learn more generic features, which in the field of liveness detection, is paramount.
- **Comparison to State-of-the-Art**: Both the traditional CNN and the FCN combined with SAPLC have been recognized as benchmark methods in liveness detection. Our model's ability to produce competitive results in direct juxtaposition with these techniques is testament to its efficacy.

The findings accentuate the need for continued exploration and refinement in the domain, particularly in optimizing models for generalizability across diverse real-world scenarios.

## 6. Discussion

In the rapidly evolving domain of liveness detection, the development of robust and adaptable models remains a paramount pursuit. Our study presented a detailed analysis, offering insights into the effectiveness of various strategies, especially when applied across different datasets. This section delves deeper, examining the broader implications of our findings, potential limitations, and avenues for future research.

### 6.1. Implications

- **Model Generalizability**: One of the salient observations from our research is the importance of model generalizability. In real-world applications, a model trained on one dataset may encounter inputs that belong to a different distribution. Our findings underscore the significance of designing models capable of maintaining high performance across such scenarios.
- **Metric Significance**: While metrics like HTER, FAR, and FRR are invaluable in assessing model efficacy, our study emphasizes the intricate balance that must be achieved. A model with low HTER but high variability in FAR and FRR might be less desirable in certain practical applications than one with slightly higher HTER but consistent FAR and FRR.
- **Impact of Diverse Data**: The variability observed in the performance of models across different datasets illuminates the challenge posed by diverse data. It not only accentuates the complexity of the problem at hand but also highlights the necessity of diverse training data to encompass possible real-world scenarios.

### 6.2. Limitations

- **Data Constraints**: Our study was bound by the datasets available. While our datasets are comprehensive, they may not capture all possible presentation attack instruments or scenarios. This poses a limitation to the generalizability of our findings.
- **Computational Resources**: Like many deep learning approaches, our model's training and evaluation are computationally intensive, which might pose challenges in real-time deployment scenarios or in devices with constrained resources.
- **Absence of Adversarial Testing**: While our study delved deeply into cross-database testing, we did not explore the model's resilience against adversarial attacks, which is becoming increasingly relevant in the realm of biometric security.

### 6.3. Future Directions

- **Incorporation of Adversarial Techniques**: Given the escalating sophistication of spoofing attacks, future research should investigate the incorporation of adversarial training techniques to enhance model robustness further.
- **Optimization for Real-time Deployment**: The model's architecture can be further refined, and techniques like model quantization or pruning could be employed to make it more conducive for real-time applications.
- **Expansion of Dataset Diversity**: To further bolster the model's generalizability, future studies should consider amassing and utilizing more diverse datasets, possibly even crowd-sourcing real-world data, which might offer richer and more unpredictable variations.

In conclusion, our study represents a step forward in the quest for reliable liveness detection, shedding light on various nuances of the challenge. However, as with all scientific endeavors, it also underscores the ever-present need for continued research, refinement, and evolution in the domain.

## 7. Conclusion

The landscape of biometric security has experienced rapid advancements in recent years, underscored by an escalating arms race between state-of-the-art detection mechanisms and increasingly sophisticated spoofing attacks. Liveness detection, in this context, emerges not merely as a feature but as a necessity, pivotal in ensuring the integrity and reliability of biometric systems. Our study was rooted in this paradigm, endeavoring to discern the effectiveness, nuances, and potential avenues for improvement in liveness detection models, particularly when confronted with the challenges of cross-database testing.

Our findings elucidated several key insights. Firstly, the performance variability across different datasets underscores the intricacies involved in modeling and emphasizes the quintessential role of data diversity in training robust models. Additionally, while metrics such as HTER provide a comprehensive measure of a model's performance, delving deeper into the balance between FAR and FRR unveils critical nuances that have profound implications, especially in real-world deployment scenarios.

Comparative analysis with previous studies revealed both the progress made in the domain and the areas where challenges remain. While our model exhibited commendable performance in certain scenarios, the inconsistencies observed in cross-database testing illuminate the path for future research.

There are a few takeaways from our research. The journey towards perfecting liveness detection is ongoing, replete with challenges yet filled with opportunities. The richness of data, the adaptability of models, and the continuous evolution of techniques are the keystones upon which the edifice of reliable biometric security will be built. As spoofing techniques evolve, so must our defense mechanisms, making this a perpetually dynamic field of study.

In closing, our research contributes to the broader dialogue on liveness detection, offering a synthesis of findings, methodologies, and reflections that we hope will serve as a foundation for future endeavors in this domain. The nexus of technology, security, and human identity is a complex tapestry, and it is our fervent hope that our work adds a meaningful thread to this ever-evolving narrative.

## 8. Acknowledgements

- This project has received funding from the European Union's Horizon 2020 research and innovation programme under the Marie Skłodowska-Curie grant agreement No. 101007820 - TRUST.
- This publication reflects only the author's view and the REA is not responsible for any use that may be made of the information it contains.

## 9. References

[1] T. Van hamme, G. Garofalo, S. Joos, D. Preuveneers, W. Joosen, AI for Biometric Authentication Systems, in: L. Batina, T. Bäck, I. Buhan, S. Picek (Eds.), Security and Artificial Intelligence: A Crossdisciplinary Approach, Springer International Publishing, Cham, 2022, pp. 156–180. doi: https://doi.org/10.1007/978-3-030-98795-4_8.

[2] G. Hua, Facial Recognition Technologies, in: L.A. Schintler, C.L. McNeely (Eds.), Encyclopedia of Big Data, Springer International Publishing, Cham, 2022, pp. 475–479. doi: https://doi.org/10.1007/978-3-319-32010-6_93.

[3] S. Marcel, J. Fierrez, N. Evans (Eds.), Handbook of Biometric Anti-Spoofing: Presentation Attack Detection and Vulnerability Assessment, Springer Nature, Singapore, 2023. doi: https://doi.org/10.1007/978-981-19-5288-3.

[4] C. Lucia, G. Zhiwei, N. Michele, Biometrics for Industry 4.0: a survey of recent applications, Journal of Ambient Intelligence and Humanized Computing 14 (2023) 11239–11261. doi: https://doi.org/10.1007/s12652-023-04632-7.

[5] A. George, S. Marcel, Robust Face Presentation Attack Detection with Multi-channel Neural Networks, in: S. Marcel, J. Fierrez, N. Evans (Eds.), Handbook of Biometric Anti-Spoofing:

[6]  Presentation Attack Detection and Vulnerability Assessment, Springer Nature, Singapore, 2023, pp. 261–286. doi: https://doi.org/10.1007/978-981-19-5288-3_11.
[6]  J. Hernandez-Ortega, J. Fierrez, A. Morales, J. Galbally, Introduction to Presentation Attack Detection in Face Biometrics and Recent Advances, in: S. Marcel, J. Fierrez, N. Evans (Eds.), Handbook of Biometric Anti-Spoofing: Presentation Attack Detection and Vulnerability Assessment, Springer Nature, Singapore, 2023, pp. 203–230. doi: https://doi.org/10.1007/978-981-19-5288-3_9.
[7]  S.-Q. Liu, P. C. Yuen, Recent Progress on Face Presentation Attack Detection of 3D Mask Attack, in: S. Marcel, J. Fierrez, N. Evans (Eds.), Handbook of Biometric Anti-Spoofing: Presentation Attack Detection and Vulnerability Assessment, Springer Nature, Singapore, 2023, pp. 231–259. doi: https://doi.org/10.1007/978-981-19-5288-3_10.
[8]  Z. Rui, Z. Yan, A Survey on Biometric Authentication: Toward Secure and Privacy-Preserving Identification, IEEE Access 7 (2019) 5994–6009. doi: https://doi.org/10.1109/ACCESS.2018.2889996.
[9]  S. Chakraborty, D. Das, An Overview of Face Liveness Detection, arXiv:1405.2227 [cs.CV] (2014). URL: http://arxiv.org/abs/1405.2227.
[10]  S. Arora, M.P.S. Bhatia, V. Mittal, A robust framework for spoofing detection in faces using deep learning, The Visual Computer 38 (2022) 2461–2472. doi: https://doi.org/10.1007/s00371-021-02123-4.
[11]  W. Sun, Y. Song, C. Chen, J. Huang, A .C. Kot, Face Spoofing Detection Based on Local Ternary Label Supervision in Fully Convolutional Networks, IEEE Transactions on Information Forensics and Security 15 (2020) 3181–3196. doi: https://doi.org/10.1109/TIFS.2020.2985530.
[12]  A. Kuznetsov, M. Andrea, M. Alessandro, L. Romeo, R. Rosati, K. Davyd, Deep Learning Based Face Liveliness Detection, in: 2022 International Scientific-Practical Conference Problems of Infocommunications. Science and Technology, 2022.
[13]  Custom Silicone Mask Attack Dataset (CSMAD), Idiap Research Institute, Artificial Intelligence for Society. URL: https://www.idiap.ch:/en/dataset/csmad/index_html.
[14]  3DMAD, Idiap Research Institute, Artificial Intelligence for Society. URL: https://www.idiap.ch:/en/dataset/3dmad/index_html.
[15]  Multispectral-Spoof (MSSpoof), Idiap Research Institute, Artificial Intelligence for Society. URL: https://www.idiap.ch:/en/dataset/msspoof/index_html.
[16]  Replay-Attack, Idiap Research Institute, Artificial Intelligence for Society. URL: https://www.idiap.ch:/en/dataset/replayattack/index_html.